\documentclass[conference]{IEEEtran}
\IEEEoverridecommandlockouts
\usepackage{cite}
\usepackage{amsmath,amssymb,amsfonts}
\usepackage{algorithmic}
\usepackage{graphicx}
\usepackage{textcomp}
\usepackage{xcolor}
\usepackage{fancyhdr}

\usepackage{subfig}
\usepackage{booktabs}
\usepackage[para,online,flushleft]{threeparttable}
\usepackage{tabularx}
\usepackage{multirow}

\makeatletter
\def\ps@IEEEtitlepagestyle{%
  \def\@oddfoot{\mycopyrightnotice}%
  \def\@evenfoot{}%
}
\def\mycopyrightnotice{%
  {\footnotesize 978-1-7281-0858-2/19/\$31.00 \textcopyright 2019 IEEE \hfill}% <--- Change here
  \gdef\mycopyrightnotice{}% just in case
}

\newcommand*\titleheader[1]{\gdef\@titleheader{#1}}
\AtBeginDocument{%
  \let\st@red@title\@title
  \def\@title{%
    \bgroup\normalfont\large\centering\@titleheader\par\egroup
    \vskip1.5em\st@red@title}
}

\def\BibTeX{{\rm B\kern-.05em{\sc i\kern-.025em b}\kern-.08em
    T\kern-.1667em\lower.7ex\hbox{E}\kern-.125emX}}
\begin{document}

\title{Deep Transfer Learning for Thermal Dynamics Modeling in Smart Buildings\\
}
\titleheader{2019 IEEE International Conference on Big Data (Big Data)}

\author{\IEEEauthorblockN{Zhanhong Jiang}
\IEEEauthorblockA{\textit{Johnson Controls} \\
\textit{507 East Michigan St.}\\
Milwaukee, USA \\
zhanhong.jiang@jci.com}
\and
\IEEEauthorblockN{Young M. Lee}
\IEEEauthorblockA{\textit{Johnson Controls} \\
\textit{507 East Michigan St.}\\
Milwaukee, USA \\
young.m.lee@jci.com}
% \and
% \and
% \and
% \and
}

\maketitle

\begin{abstract}
Thermal dynamics modeling has been a critical issue in building heating, ventilation, and air-conditioning (HVAC) systems, which can significantly affect the control and maintenance strategies. Due to the uniqueness of each specific building, traditional thermal dynamics modeling approaches heavily depending on physics knowledge cannot generalize well. This study proposes a deep supervised domain adaptation (DSDA) method for thermal dynamics modeling of building indoor temperature evolution and energy consumption. A long short term memory network based Sequence to Sequence scheme is pre-trained based on a large amount of data collected from a building and then adapted to another building which has a limited amount of data by applying the model fine-tuning. We use four publicly available datasets: SML and AHU for temperature evolution, long-term datasets from two different commercial buildings, termed as Building 1 and Building 2 for energy consumption. We show that the deep supervised domain adaptation is effective to adapt the pre-trained model from one building to another building and has better predictive performance than learning from scratch with only a limited amount of data. 
\end{abstract}


\section{Introduction}
Modern buildings have been of crucial importance in daily life for human's activities and can significantly affect the productivity, and physiological and psychological characteristics~\cite{enescu2017review}. One critical function, maintaining occupant thermal comfort in buildings, especially large-scale commercial buildings, heavily depends on the efficacy and efficiency of control and maintenance strategies. To design feasible control strategies, it typically requires sufficiently adaptive environment models to capture the dynamics of thermal zones in buildings and to adjust for the unknown disturbances, either being internal or external. Recently data-driven methodologies have been receiving considerable attention and shown to be effective in capturing the thermal dynamics, such as artificial neural networks~\cite{reynolds2018zone}, support vector machine~\cite{molina2017data}, autoregressive models~\cite{jiang2018data,chinde2015comparative,lee2015optimal}, and deep neural networks~\cite{mocanu2016deep}.

While a data-driven model can be adopted easily to describe the thermal dynamics for a specific building in a time interval, there exists difficulty that needs to be addressed on transferring established models to more buildings. Generally speaking, for different buildings or different components within the same building, one may repeatedly derive data-driven models using different data and ignoring already existing models. However, deriving from scratch every time based on different data can be time-consuming and even unfeasible without enough historical data, especially when one building is brand-new and not yet commissioned. Therefore, transferability is quite critical in the building thermal dynamics modeling and has not been sufficiently explored previously. \textit{Transfer Learning} has received considerable attention recently and been successfully used in various areas, e.g., indoor localization~\cite{pan2008transfer}, image-processing~\cite{motiian2017unified}, natural language processing~\cite{conneau2017supervised}, and biological applications~\cite{muandet2013domain}. If a model trained in one domain (any variable of interest with a large amount of data) could be adapted in another domain (any variable of interest with a limited amount of data), then one is able to avoid training a model from scratch and some valuable prior knowledge could be transferred accordingly for improving the model learning performance. While autoregressive models are able to represent well the building energy system thermal dynamics, it could be difficult to determine what order of model should be selected properly. Although other aforementioned machine learning methods were shown to be useful for approximating the thermal dynamics, the stochastic nature of real buildings resulting in complex data patterns requires more advanced models. Hence, the model adopted in this paper is a category of deep neural networks (DNN), which have been shown provably efficient for time-series prediction~\cite{ding2015deep,zhao2017lstm}.  

In this context, we propose a deep supervised domain adaptation (DSDA) method for thermal dynamics modeling of building indoor temperature evolution and energy consumption using a deep learning model, Long Short Term Memory Network based Sequence to Sequence (LSTM S2S)~\cite{sutskever2014sequence} scheme. We pre-train the LSTM S2S model using a large amount of data from one building (referred to as source building). Then, we adopt parameters from the pre-trained model to initialize parameters of a model defined for another building (referred as to target building). We use a limited amount of data from the target building to fine-tune the model parameters.
To the best of our knowledge, this is the first attempt to address the knowledge transfer using deep transfer learning techniques in building thermal dynamics. We show that the proposed approach outperforms learning from scratch given only a limited amount of data. Our experimental results also imply that useful knowledge can be transferred between different tasks for improving performance. 

One key concerning in this paper is that the resultant adaptation is between two close domains. Essentially, the source and target buildings may be required to have similar measured variables and parameters. This concerning can be practically justified as for most buildings, key variables and parameters are quite similar, although sample distributions can be significantly different due to various geographic locations or operation conditions. Nonetheless, our experimental results suggest the sampling frequency between two different domains is not required to be the same for the DSDA, which is practically useful in data collection in real buildings. 

\textbf{Related Work.} Grubinger el al.~\cite{grubinger2017generalized} developed a generalized online transfer learning for climate control in residential buildings and showed promising results on the convergence and experiments. They also used Transfer Component Analysis~\cite{pan2011domain} to allow several sources instead of single source to benefit the target task. Although such a framework enables the transferability between different houses, it relies on the physics-based modeling approach, which can be quite complicated and computationally intractable. In building design, Singaravel et al.~\cite{singaravel2018deep} came up with the component-based machine learning (CBML) to incorporate transfer learning as an approach to predict the cooling and heating energy with high accuracy. In another work~\cite{geyer2018component}, they applied the similar method to conduct the parameterized components learning of the design as well. Mocuna et al.~\cite{mocanu2016unsupervised} also developed a cross-building transfer learning framework for unsupervised energy prediction in a smart grid context. 
\section{Preliminaries}
In this context, we consider two datasets which correspond to the source ($S$) and target ($T$) buildings and denote by $\mathcal{D}_S$ and $\mathcal{D}_T$, respectively. Then, we have $\mathcal{D}_S=\{(\mathbf{x}^S_i, \mathbf{y}^S_i)\}^m_{i=1}, \mathcal{D}_T=\{(\mathbf{x}^T_j, \mathbf{y}^T_j)\}^n_{j=1}$
where $\mathbf{x}$ which is a realization from a random variable $X$, represents either a source or target domain input in $\mathcal{X}$, and $\mathbf{y}$ which is a realization from a random variable $Y$, represents either a source or target domain output in $\mathcal{Y}$. Since we focus on time series prediction, specifically, for the source building, $\mathbf{x}^S$ is defined as $x_1^Sx_2^S...x_K^S$ to be a time series of length $K$, where $x_k^S\in\mathbb{R}^d, k\in\{1,2,...,K\}$ represents a $d$ variables of vector at the time-instant $k$. Similarly, we have $\mathbf{y}^S=y_1^Sy_2^S...y_L^S$, where $y_l^S\in\mathbb{R}^p, l\in\{1,2,...,L\}$ indicates a $p$ variables of vector at the time-instant $t$. By following the same definitions, for the target domain, $\mathbf{x}^T = x_1^Tx_2^T...x_O^T$, where $x^T_o\in\mathbb{R}^d, o\in\{1,2,...,O\}$, and $\mathbf{y}^T = y_1^Ty_2^T...y_U^T$, where $y^T_u\in\mathbb{R}^p, h\in\{1,2,...,H\}$. One motivation for us to use transfer learning is that $n\ll m$, which implies that the number of samples in the target dataset is much smaller than that in the source dataset.

% We illustrate the specific problem setup for the thermal dynamic modeling in smart buildings. 
One quite practical issue for thermal dynamics modeling using data-driven techniques is that for some brand-new buildings which are not commissioned yet, modeling cannot be conducted with only a limited amount of data, especially for deep learning models. Therefore, domain adaptation can leverage the transferrability from one building to another building.
We consider \textit {supervised domain adaptation}, which typically refers to different domains resulting in a covariate shift \cite{bickel2009discriminative} between $X^S$ and $X^T$, and the model is pre-trained using $\mathcal{D}_S$ via the supervised learning way in our problem. 
% and Fig.~\ref{Illustrative_example} is used for the purpose of discussion. 
% \begin{figure}[h!]
%   \includegraphics[width=0.4\textwidth]{Figures/Illustrative_example}
%     \centering
%   \caption{Illustrative example: Old building has rich data to well train a model; brand-new building has a limited amount of data which may not train well the model; the model from old building can be a pre-trained model which transfers its weights to the brand-new building to avoid training from scratch}
%   \label{Illustrative_example}
% \end{figure}
Under this setting, the goal is to learn a prediction function mapping from $\mathcal{X}$ to $\mathcal{Y}$ that is able to perform well on $\mathcal{D}_T$.

\section{DSDA: from source building to target building}
\begin{figure*}[h!]
  \includegraphics[width=0.7\textwidth]{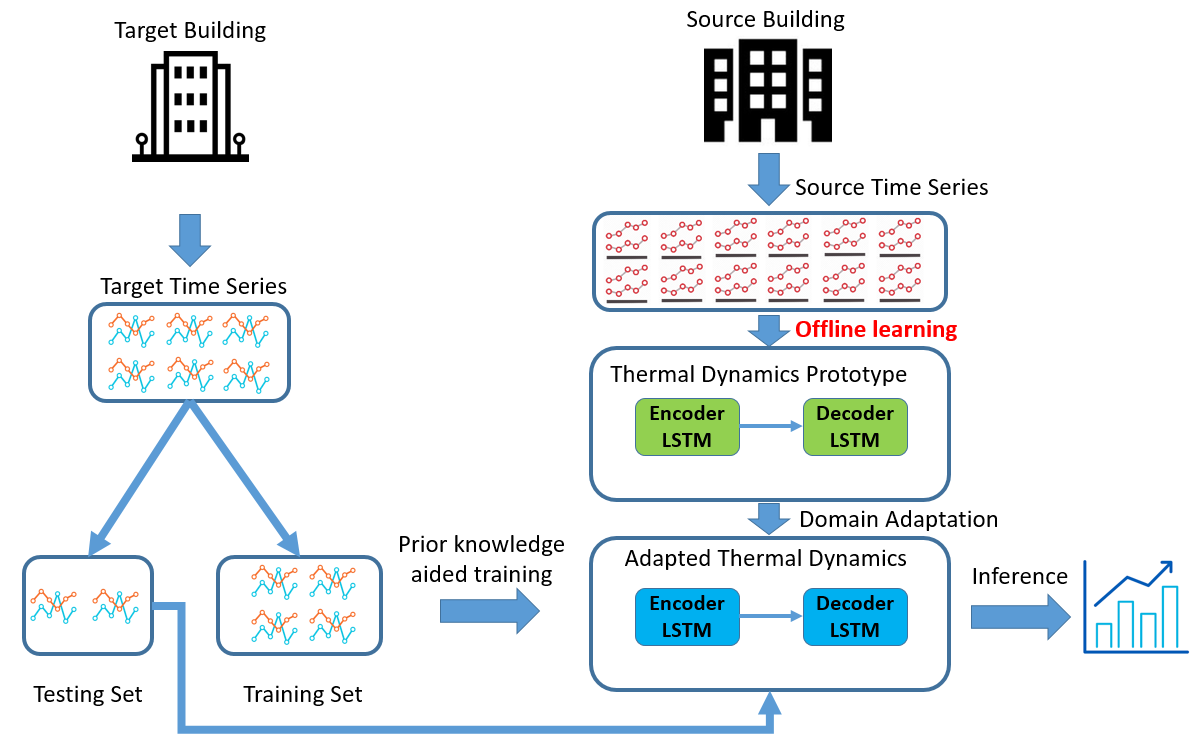}
    \centering
  \caption{Proposed DSDA: LSTM S2S model pre-trained and adapted between source and target buildings}
  \label{DSDA}
\end{figure*}

This section presents how DSDA can be applied to the thermal dynamics modeling with building time series by means of a deep model, i.e., Long Short Term Memory Network based Sequence to Sequence (LSTM S2S) model. Taking the zone temperature evoluation (or energy consumption) as a time series prediction task, we train LSTM S2S as a deep regressor using a large dataset from the source building and then adapt it to the target building involving different but related tasks. Specifically, we consider the following approach to adapt the pre-trained model to an unseen related target task, as shown in Fig.~\ref{DSDA}. We first collect data from the source building to do the off-line learning by pre-training the LSTM S2S model; we then use the pre-trained model to initialize the parameters of a model for the target building with some unseen tasks, and fine-tune the model for the target building using a limited amount of data, which is a prior-knowledge aided training. Although some similar ideas were proposed to solve image classification problems~\cite{motiian2017unified}, few results have been reported in time series prediction, in particular for building thermal dynamics modeling.
To get a limited amount of data for the target building, one can conduct simple system identification experiments. However, due to different outside environment conditions, some variables of interest can be significantly different at different times. In that case, the task adaptation technique can still be applied because periodically retraining the model maintains a certain level of accuracy by incorporating more information of the outside environment condition.
\begin{figure*}[h!]
  \includegraphics[width=0.5\textwidth]{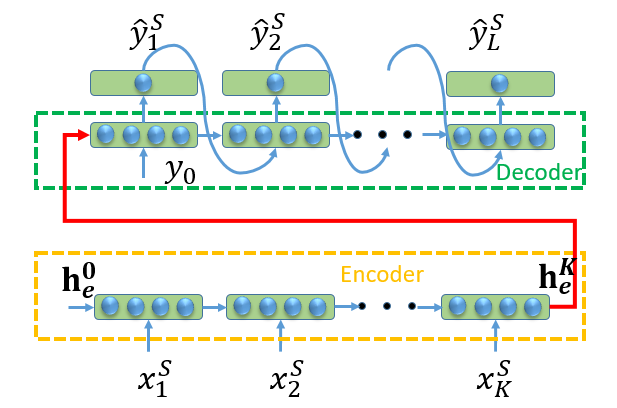}
    \centering
  \caption{Schematic of LSTM S2S Model}
  \label{LSTM S2S}
\end{figure*}
\subsection{Pre-training LSTM S2S}
%LSTM S2S model has been 
% widely been used for machine translation~\cite{sutskever2014sequence} and video analytics~\cite{venugopalan2015sequence}. 
% recently applied to multi-sensor anomaly detection~\cite{malhotra2016lstm} and building energy load forecast~\cite{marino2016building} and shows fairly good performance.
Due to the space limit, we skip the updates for each LSTM cell and refer readers to~\cite{park2018sequence} for more details. For the pre-trained model, the LSTM S2S architecture is adopted accordingly, which maps input $\mathbf{x}_i^S\in\mathcal{D}_S$ to output $\mathbf{\hat{y}}_i^S$, as shown in Fig.~\ref{LSTM S2S}. Specifically, each input $\mathbf{x}_i^S$ is encoded as a vector, which corresponds to the final state of the encoder denoted by $\mathbf{h}^{K}_e\in\mathbb{R}^c$, where $c$ is the number of LSTM units in the hidden layer of the encoder. Then, $\mathbf{h}^{K}_e$ is used as the initial state for activating the decoder with the current measurement $y_{0}$, on top of which a dense layer with linear activation is used to recursively predict each time step in $\mathbf{y}_i^S$. In every update, the decoder feeds the predicted output $\hat{y}_l$ obtained from the previous update to the input for the current update.
It should be noted that one can either apply the teacher forcing or non-teacher forcing way ~\cite{lamb2016professor} to train the LSTM S2S architecture. For avoiding the learning becoming ``lazy", which means the corresponding predicted output is obtained by making small modification to the corresponding ground truth, non-teacher forcing is adopted. 
The parameters of the network are obtained by minimizing the mean square error loss given $\mathcal{J}_1$, i.e., $\mathcal{J}_1(y_i^l,\hat{y}_i^l) = \frac{1}{n\times L}\sum_{i=1}^n\sum_{l=1}^L|y_i^l-\hat{y}_i^l|^2$, via some adaptive gradient descent methods. $\hat{y}_l$ is the predicted output at time instant $l$. For denotation, we move $l$ to the superscript in $\mathcal{J}_1$. During training, the decoder can directly uses the ground truth $y^l_i$ as input instead of $\hat{y}^l_i$, which can speed up the training.
% \begin{equation}\label{encoder}
% \mathbf{h}^k_e = f_e(\mathbf{W}_{hh,S}\mathbf{h}^{k-1}_e+\mathbf{W}_{hx,S}x_{k-1})
% \end{equation}
% \begin{equation}\label{decoder}
% \mathbf{h}^l_d = f_d(\mathbf{W}'_{hh,S}\mathbf{h}^{l-1}_d)
% \end{equation}
% \begin{equation}\label{dense}
% \hat{y}_l = g_d(\mathbf{W}'_{yh,S}\mathbf{h}^{l}_d + \mathbf{b}_{d,S})
% \end{equation}
%\begin{equation}\label{mse}
%\end{equation}
% Eqs.~\ref{encoder}, ~\ref{decoder}, and~\ref{dense} respectively indicate the \textit{nonlinear state-space} representations of encoder, decoder, and dense layers
% Eq.~\ref{mse} represents the mean square error between the ground truth and predicted output. 
% $f_e$ and $f_d$ are the nonlinear functions respectively for the encoder and decoder, $\mathbf{W}_{hh,S}, \mathbf{W}_{hx,S}, \mathbf{W}'_{hh,S}, \mathbf{W}'_{yh,S}, \mathbf{b}_{d,S}$ are the parameters for encoder, decoder and dense layers. 
% $g_d$ representing the activation function between the LSTM layer and dense layer is given without loss of generality, although in this context it is linear. 
\subsection{Fine-tuning LSTM S2S}
After pre-training the LSTM S2S model using a larget amount of data from the source building, we adapt the model to the target building for initializing the target task specific LSTM parameters. Different from classification, which may require to freeze parameters of the encoder and decoder and only to fine-tune the parameters of the dense layer, we re-train the whole model using a limited amount of data from the target building in this context. Compared to typical convolutional neural networks using multiple layers to extract features, LSTM S2S can be treated as a nonlinear state-space model which holds the recurrency to model the temporal dynamics and single layer for either the encoder and decoder is practically feasible.
% We denote by $\mathbf{W}_{hh,T}, \mathbf{W}_{hx,T}, \mathbf{W}'_{hh,T}, \mathbf{W}'_{yh,T}, \mathbf{b}_{d,T}$ new parameters of the adapted model. Therefore, the following time series prediction task is obtained: 
% \begin{equation}\label{adaptation}
%     \tilde{y}_u=g_d(\mathbf{W}'_{yh,T}\mathbf{h}_d^u+\mathbf{b}_{d,T})
% \end{equation}
Hence, the loss given $\mathcal{J}_2$ can be immediately obtained:
%\vspace{-0.2cm}
$\mathcal{J}_2(y_j^u,\tilde{y}_j^u) = \frac{1}{m\times U}\sum_{j=1}^m\sum_{u=1}^U|y_j^u-\tilde{y}_j^u|^2$. By minimizing $\mathcal{J}_2$ with a limited amount of data, the adapted model for the target building is correspondingly acquired and then can be used for the inference.
% \vspace{-0.4cm}
\begin{table}[h]
\caption{Description of Datasets}
\begin{center}
\begin{threeparttable}
\begin{tabular}{c c c c c}
    \toprule
    \textbf{Dataset} & \textbf{SF\tnote{1}} & \textbf{Size} &\textbf{\# of features}& \textbf{Domain} \\ \midrule
      SML  & 15 min                    & 1373    & 15 &Target \\
      AHU  & 1 min                     &  35098                   &15& Source  \\\midrule
      Building 1& 15 min & 2000 &4& Target\\
      Building 2& 15 min&  34940 &4& Source\\
      \bottomrule
\end{tabular}
\begin{tablenotes}
\item[1] Sampling frequency
\end{tablenotes}
\end{threeparttable}
\end{center}
\label{table:dataset}
\end{table}
\begin{table}[h]
\caption{Comparison of Metrics between DSDA and learning from scratch for testing (temperature evolution)} 
\begin{center}
\begin{threeparttable}
\begin{tabular}{c c c c c}
    \toprule
    \textbf{Dataset} & \textbf{CVRMSE} & \textbf{NMBE} & \textbf{MAPE} &\textbf{RMSE} \\ \midrule
      SML(15 min)&5.983\%  &-5.099\%  &5.253\% &1.274\\
      AHU$\to$SML(15 min)   & \textbf{1.671\%}                   & \textbf{-0.877\%}                   &\textbf{1.396\%} &\textbf{0.355}\\ \midrule
      SML(2 h)&8.421\%&-7.442\%&7.612\%&1.788\\
      AHU$\to$SML(2 h)&\textbf{3.674\%}&\textbf{-2.721\%}&\textbf{2.945\%}&\textbf{0.780}\\\midrule
      SML(4 h)&10.613\%&-9.534\%&9.747\%&2.243\\
      AHU$\to$SML(4 h)&\textbf{7.513\%}&\textbf{-5.381\%}&\textbf{6.055\%}&\textbf{1.588}\\\midrule
      SML(6 h)&12.405\%&-11.099\%&11.233\%&2.607\\
      AHU$\to$SML(6 h)&\textbf{11.143\%}&\textbf{-7.198\%}&\textbf{8.618\%}&\textbf{2.342}\\
    %   Building 1&5.715\% & \textbf{-0.930\%}  &4.308\% & 13.198\\
    %   Buildings 2$\to$ 1   &\textbf{4.842\%}                    & -1.647\%                    & \textbf{3.402\%}  & \textbf{11.182}\\ %\midrule
    %   AHU$\to$ Building 1 &5.699\% &-0.982\% &3.937\%&13.160\\
    %   Building 2$\to$ SML &6.918\% &-3.925\% &6.614\%&1.454 \\ 
      \bottomrule
\end{tabular}
% \begin{tablenotes}
% \item[1] Sampling frequency
% \end{tablenotes}
\end{threeparttable}
\end{center}
\label{table:metrics}
\end{table}

\begin{table}[h]
\caption{Comparison of Metrics between DSDA and learning from scratch for testing (energy consumption)} 
\begin{center}
\begin{threeparttable}
\begin{tabular}{c c c c c}
    \toprule
    \textbf{Dataset} & \textbf{CVRMSE} & \textbf{NMBE} & \textbf{MAPE} &\textbf{RMSE} \\ \midrule
      Building 1(15 min)&5.715\%  &\textbf{-0.930\%}  &4.308\% &13.198\\
      Building2$\to$1(15 min)   & \textbf{4.842\%}                   & -1.647\%                   &\textbf{3.402\%} &\textbf{11.182}\\ \midrule
      Building 1(2 h)&7.016\% & -2.713\%  &4.699\% & 16.220\\
      Building2$\to$1(2 h)   &\textbf{5.497\%}          & \textbf{-2.334\%}& \textbf{3.672\%}  & \textbf{12.709}\\\midrule
      Building 1(4 h)&8.033\% & \textbf{-1.835\%}  &4.871\% & 18.709\\
      Building2$\to$1(4 h)   &\textbf{6.466\%}          & -2.717\%& \textbf{4.241\%}  & \textbf{15.058}\\\midrule
      Building 1(6 h)&11.344\% & \textbf{-0.913\%}  &6.093\% & 26.600\\
      Building2$\to$1(6 h)   &\textbf{7.099\%}          & -3.208\%& \textbf{4.763\%}  & \textbf{16.646}\\
    %   Building 2$\to$ SML &6.918\% &-3.925\% &6.614\%&1.454 \\ 
      \bottomrule
\end{tabular}
% \begin{tablenotes}
% \item[1] Sampling frequency
% \end{tablenotes}
\end{threeparttable}
\end{center}
\label{table:metrics_1}
\end{table}
\begin{table}[h]
\caption{Comparison of RMSE between cross-task and learning from scratch for testing} 
\begin{center}
\begin{threeparttable}
\begin{tabular}{c c c c c}
    \toprule
    \textbf{Dataset} & \textbf{15 min} & \textbf{2 h} & \textbf{4 h} &\textbf{6 h} \\ \midrule
      SML&1.274  &1.788  &2.243 &2.607\\
      Building2$\to$SML   & \textbf{0.403}                   & \textbf{0.735}                   &\textbf{1.016} &\textbf{1.454}\\ \midrule
      Building 1&13.198 & 16.220  &18.709 & 26.600\\
      AHU$\to$Building 1   &\textbf{13.160} & \textbf{15.717}& \textbf{16.492}  & \textbf{19.457}\\
    %   Building 1(4 h)&8.033\% & \textbf{-1.835\%}  &4.871\% & 18.709\\
    %   Building2$\to$1(4 h)   &\textbf{6.466\%}          & -2.717\%& \textbf{4.241\%}  & \textbf{15.058}\\\midrule
    %   Building 1(6 h)&11.344\% & \textbf{-0.913\%}  &6.093\% & 26.600\\
    %   Building2$\to$1(6 h)   &\textbf{7.099\%}          & -3.208\%& \textbf{4.763\%}  & \textbf{16.646}\\
    %   Building 2$\to$ SML &6.918\% &-3.925\% &6.614\%&1.454 \\ 
      \bottomrule
\end{tabular}
% \begin{tablenotes}
% \item[1] Sampling frequency
% \end{tablenotes}
\end{threeparttable}
\end{center}
\label{table:RMSE}
\end{table}
\section{Experiments}
This paper considers four publicly available datasets: SML~\cite{zamora2014line} and AHU~\cite{OpenEI_1} for builidng indoor temperature evolution; two long-term datasets from two different commercial buildings for energy consumption~\cite{OpenEI}. Please see the Table~\ref{table:dataset} for more details. SML and AHU respectively have 15 different feature inputs and the similar output of interest is indoor temperature, while Buildings 1 and 2 have 4 identical feature inputs (total power consumption, outdoor temperature, day of week, time of day) and the common output of interest is the whole building energy consumption. For completeness, the appendix includes feature details of AHU and SML datasets. For implementation, we use the whole source dataset to pre-train the model, and then split the target dataset into training and testing in chronological time with a ratio being 0.67. The ratio is fixed in this context, while a minimum ratio will be in the future work to figure out the minimum amount of data needed for DSDA. We perform 15 min, 2 hours, 4 hours, and 6 hours ahead prediction and it should be noted that the prediction horizon depends on the target tasks.
% \begin{figure}
% \centering
% \begin{subfigure}
% \centering
%     \includegraphics[width=0.3\textwidth]{Figures/SML_15_min}
%     %\par\medskip
%     %\vspace{-0.2cm}
%     \includegraphics[width=0.3\textwidth]{Figures/AHU_SML_15_min.png}
% \end{subfigure}\vspace{-0.2cm}
% \caption{15 min ahead prediction for temperature: (Top) Learning from scratch with SML data; (Bottom) DSDA: model pre-trained by AHU data and adapted to SML data}
% \label{temperature_evolution}
% \end{figure}

% \begin{figure}
% \centering
% \begin{subfigure}
% \centering
%     \includegraphics[width=0.3\textwidth]{Figures/Building_1_15_min}
%     %\par\medskip
%     %\vspace{-0.2cm}
%     \includegraphics[width=0.3\textwidth]{Figures/Building_2_to_1_15_min.png}
% \end{subfigure}\vspace{-0.2cm}
% \caption{15 min ahead prediction for energy: (Top) Learning from scratch with Building 1 data; (Bottom) DSDA: model pre-trained by Building 2 data and adapted to Building 1 data}
% \label{energy_consumption}
% \end{figure}
The architecture of LSTM S2S has one LSTM layer for both the encoder and decoder, respectively. To pre-train LSTM S2S using the data from the source building and to fine-tune the parameters using the data from the target building, we adopt the mini-batch stochastic optimization based on Adam optimizer~\cite{kingma2014adam}. We also set the hyperparameters based on the optimal performance we obtain. Tables~\ref{table:metrics} and~\ref{table:metrics_1} show the metrics of DSDA used for this study and the comparison with learning from scratch. They are the coefficient of variance of root mean square error (CVRMSE), normalized mean bias error (NMBE), mean absolute percentage error (MAPE), and root mean square error (RMSE).
%~\cite{coakley2012calibration}.
%\subsection{Case Study 1: temperature evolution}

\textbf{Case Study 1: temperature evolution.} As Table~\ref{table:dataset} shows, we use two datasets, i.e., SML and AHU for validating the domain adapatation of temperature evolution. Prediction tasks in the source and target domains are quite related, as both of them are indoor temperature evolution, even with different features. From Table~\ref{table:metrics}, which shows predictions of different horizons, it can be observed that with pre-trained model by the AHU data, the predictive performance can be improved in terms of each metric for the SML data. To study the transferability between different tasks, Table \ref{table:RMSE} shows the comparison between cross-task and learning from scratch. We pre-train the model using the Building 2 data to extract temporal dependencies for the energy consumption and then adapt it to learn the temperature evolution for the target task. It should be noted that the number of input features of SML is adjusted such that the prediction task can be conducted accordingly. From Table~\ref{table:RMSE}, the results of Building 2$\to$ SML show that even if the pre-trained model is from another different source which has a completely different task, temporal dependencies of the source data can still be extracted to improve the predictive performance in the target task, compared to learning from scratch. 
% Another implication can be made from Tables~\ref{table:metrics} is that for DSDA, sampling frequency is not required the same in both of the source and target buildings when learning from AHU to SML, which can be practically useful in different buildings.

%\subsection{Case Study 2: energy consumption}
\textbf{Case Study 2: energy consumption.} In this case study, we use the total energy consumption data from two different commercial buildings to validate the proposed scheme. Prediction tasks for the source and target domains are the same in this case with identical feature inputs, as mentioned above. As shown in Table~\ref{table:metrics_1}, while DSDA and learning from scratch have close performance, the former one still outperforms by using the pre-trained model. The only slight improvement is attributed to that in the total energy consumption, a single good pattern can be quickly extracted even with a small size of data. It is noted that NMBE in this case does not indicate improvement since based on the definition of NMBE, the error cancellation may take place for the data. Table~\ref{table:RMSE} also shows the prediction of energy consumption based on the trained model using temperature features, i.e., AHU$\to$Building 1. Similarly, the trasnferability across different tasks is useful for improving performance, which is consistent with the conclusion made in Case Study 1.
%\subsection{Case Study 3: temperature $\to$ energy}
% \textbf{Case Study 3: temperature $\to$ energy.} We pre-train the model using the AHU data to extract the temporal dependencies for the temperature evolution and then adapt it to learn the energy consumption by initializing the model parameters for the target task. It should be noted that the number of input features of AHU is adjusted such that the prediction task for Building 1 can be conducted accordingly. From Table~\ref{table:metrics}, the results of AHU$\to$ Building 1 show that even if the pre-trained model is from another different source which has a completely different source task, temporal dependencies of the source data can still be extracted to assist in improving the predictive performance in the target task, compared to learning from scratch. While Buildings 2$\to$ 1 overall outperforms AHU$\to$ Building 1, learning from multiple sources is out of the scope in this paper and it will be the future work.
\section{Conclusion and Future Works}
This study proposed a deep supervised domain adaptation for thermal dynamics modeling in smart buildings. It aims at solving the problem of generalizing an established model from one building to another building. We adpat a pre-trained LSTM S2S model from the source building to the target building using the model fine-turning method. Extensive numerical results show that such a proposed scheme outperforms learning from scratch. The approach is critically important in scenarios when buildings don't have enough data for learning a model and allows facility managers to quickly establish maintenance and control strategies for building energy systems.
Future directions include how to learn from multiple sources to single target as well as unsupervised domain adaptation for missing variables.

\bibliographystyle{IEEEtran}
\bibliography{dsda}

% \bibitem{b1} G. Eason, B. Noble, and I. N. Sneddon, ``On certain integrals of Lipschitz-Hankel type involving products of Bessel functions,'' Phil. Trans. Roy. Soc. London, vol. A247, pp. 529--551, April 1955.
% \bibitem{b2} J. Clerk Maxwell, A Treatise on Electricity and Magnetism, 3rd ed., vol. 2. Oxford: Clarendon, 1892, pp.68--73.
% \bibitem{b3} I. S. Jacobs and C. P. Bean, ``Fine particles, thin films and exchange anisotropy,'' in Magnetism, vol. III, G. T. Rado and H. Suhl, Eds. New York: Academic, 1963, pp. 271--350.
% \bibitem{b4} K. Elissa, ``Title of paper if known,'' unpublished.
% \bibitem{b5} R. Nicole, ``Title of paper with only first word capitalized,'' J. Name Stand. Abbrev., in press.
% \bibitem{b6} Y. Yorozu, M. Hirano, K. Oka, and Y. Tagawa, ``Electron spectroscopy studies on magneto-optical media and plastic substrate interface,'' IEEE Transl. J. Magn. Japan, vol. 2, pp. 740--741, August 1987 [Digests 9th Annual Conf. Magnetics Japan, p. 301, 1982].
% \bibitem{b7} M. Young, The Technical Writer's Handbook. Mill Valley, CA: University Science, 1989.
% \end{thebibliography}
\section*{Appendix}
SML: indoor temperature, carbon dioxide, relative humidity, lighting, rain, sun dusk, wind, sun light in east/west/south facade, sun irradiance, outdoor temperature, outdoor relative humidity, day of week, time of day\\
AHU: indoor temperature, indoor temperature setpoint, supply fan command, outdoor temperature, return air temperature, mixed air temperature, outside air damper command, discharge air temperature setpoint, discharge air temperature, supply fan speed command, discharge air static pressure, reutrn fan command, return fan speed command, day of week, time of day

\end{document}